\documentclass[letterpaper, 10 pt, conference]{ieeeconf}  

\IEEEoverridecommandlockouts                              

\usepackage{graphicx}
\usepackage{amsmath}
\usepackage{amssymb}
\usepackage{booktabs}
\usepackage{hyperref}
\usepackage{amssymb}
\usepackage{amsfonts}
\usepackage{blindtext}

\usepackage{times}
\usepackage{epsfig}
\usepackage{graphicx}
\usepackage{amsmath}
\usepackage{amssymb}

\usepackage{algorithm}
\usepackage{algpseudocode}

\usepackage{lipsum}
\usepackage{graphics} 
\usepackage{epsfig} 
\usepackage{mathptmx} 
\usepackage{times} 
\usepackage[normalem]{ulem}
\usepackage{subcaption}
\usepackage[normalem]{ulem}
\usepackage{todonotes}
\usepackage{setspace}
\usepackage{mathtools}

\usepackage{tabularx} 
\usepackage{makecell}

\hypersetup{
    colorlinks=true,
    urlcolor=magenta,
}

\title{\LARGE \bf
Multi-Vehicle Trajectory Prediction at Intersections using State and Intention Information
}

\author{Dekai Zhu$^{1*}$, Qadeer Khan$^{1*}$ and Daniel Cremers$^{1}$
\thanks{$^{1}$ Computer Vision Group
, CIT, Technical University of Munich. }
\thanks{This work was funded by the Munich Center for Machine Learning}
\thanks{* These authors contributed equally}
}

\providecommand{\keywords}[1]{\textbf{\textit{Index terms---}} #1}

\begin{document}

\maketitle
\thispagestyle{empty}
\pagestyle{empty}


\begin{abstract}
Traditional approaches to prediction of future trajectory of road agents rely on knowing information about their past trajectory. This work rather relies only on having knowledge of the current state and intended direction to make predictions for multiple vehicles at intersections. Furthermore, message passing of this information between the vehicles provides each one of them a more holistic overview of the environment allowing for a more informed prediction. This is done by training a neural network which takes the state and intent of the multiple vehicles to predict their future trajectory.  Using the intention as an input allows our approach to be extended to additionally control the multiple vehicles to drive towards desired paths. Experimental results demonstrate the robustness of our approach both in terms of trajectory prediction and vehicle control at intersections. The complete  training and evaluation code for this work is available here: \url{https://github.com/Dekai21/Multi_Agent_Intersection}.
\end{abstract}

\section{Introduction}\label{sec:introduction}
Over the past decade, deep learning has made tremendous strides towards the ultimate goal of achieving full driving autonomy \cite{sae}. Self-driving vehicles deploy a suite of different sensors such as RADAR, GPS, IMU, LIDAR, cameras or their combination for various tasks such as object detection, classification, localization and navigation \cite{7960151,8832695,8836794,8443497,8806152}. Among them, vision based sensors (Cameras, Lidar etc.) have been demonstrated to be most promising in achieving at par human driving performance. This is because  these sensors are closest to emulating the traits of human vision in perceiving the driving environment. Coupled with other sensors, they have been successful in various tasks such as emergency braking \cite{6704164,7457666}, lane keeping \cite{9518365,8612566}, pedestrian detection, object tracking \cite{kim2021eagermot,meinhardt2021trackformer} etc. However, such line-of-sight sensors mounted on ego-vehicles are primarily concerned with tasks involving single vehicles and therefore have several limitations of their own: 

\begin{itemize}
    \item They can only partially observe an environment due to limited field of view, occlusions etc. Hence, they may not be feasible for executing maneuvers at hustling areas such as traffic intersections. This is important since a sizable fraction of vehicle collisions occur at traffic intersections \cite{MARZOUG2022126599} which also tend to be more severe \cite{chang2003relationship}. 
    \item Each vehicle has an independent sensor and a separate processing setup. Therefore, the combined computational power needed for all the vehicles would be high. Moreover, these resources occupy space within the ego-vehicle and may even require cooling. 
    \item Simulated engines have played a crucial role in testing and evaluating autonomous driving algorithms. However, sensor data such as images rendered in simulation may not be  a true reflection of reality. Hence, this domain shift  would preclude deployment in the real world.
\end{itemize}

\begin{figure*}
    \centering
    \includegraphics[width=1.75\columnwidth]{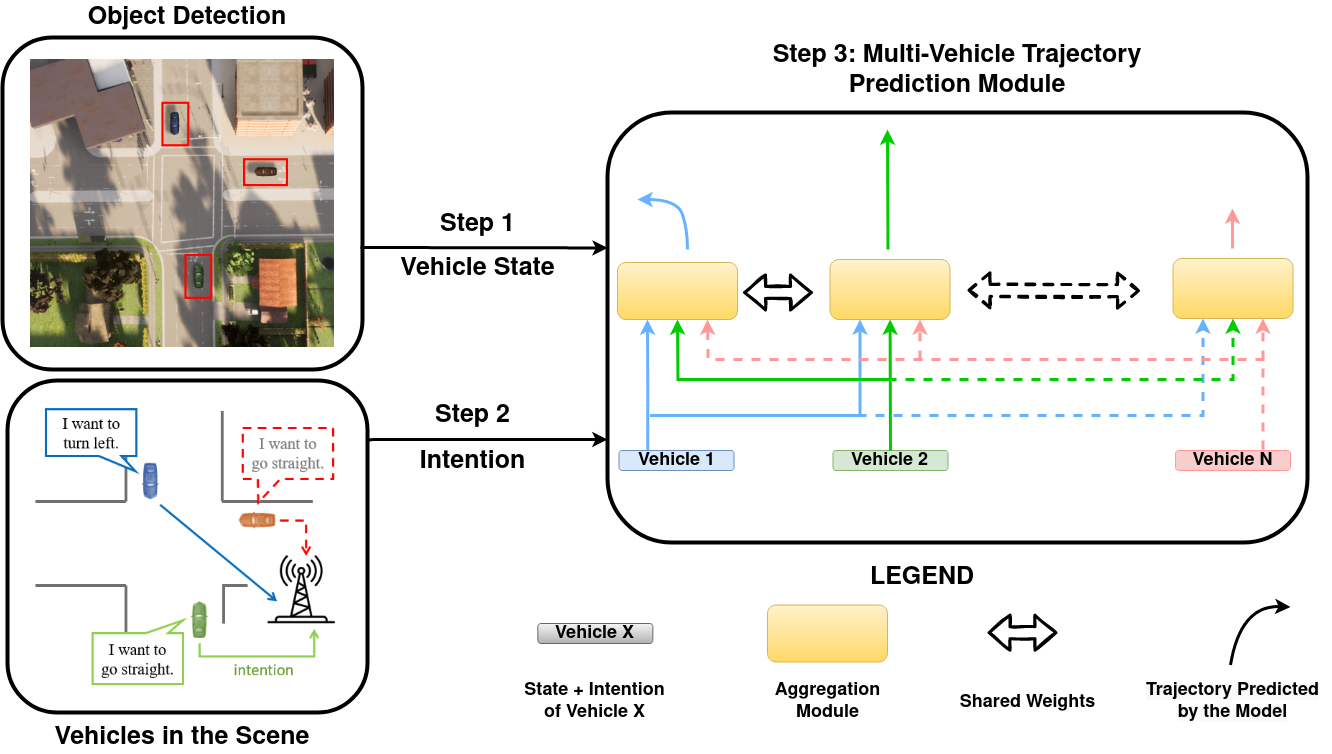}
    \caption{\textbf{Multi-vehicle Trajectory Prediction Framework:} \textit{Step 1:} Object detection is done on a top-down image of an intersection to extract out the state information of each vehicle in the scene. This information is sent to the Multi-vehicle Trajectory Prediction (MTP) module. \textit{Step 2:} Meanwhile, each vehicle in the scene also sends its intention to the MTP. The intention information informs the MTP whether a certain vehicle intends to turn left, right or keep going straight at the upcoming intersection. \textit{Step 3:} The MTP then passes this combined state and intention information to the aggregation sub-module to predict the future trajectory for each vehicle. Note that the trajectory prediction for each vehicle is not only dependent on its own state and intention information but also considers that of other vehicles too. The focus of this work is on the MTP module, where we show how the future trajectory of multiple vehicles can be predicted simultaneously using their state and intent information. }
    \label{fig:overview}
    \vspace{-3mm}
\end{figure*}

To overcome the issue associated with partial observability at critical areas such as intersections, a camera can be deployed in a Birds-Eye-View (BEV) manner simultaneously observing all agents in the scene as depicted in Figure \ref{fig:overview}. Such top-down images are commonplace for trajectory prediction \cite{978-3-319-46484-8_33,inproceedingsTopview}. The state of the agents can be captured with a camera permanently mounted on a high infrastructure \cite{Lerner2007CrowdsBE,5459260} or using drone imagery with up to centimeter accuracy precision \cite{9304839} using vision based object detection algorithms. This state information for each vehicle can then be used to predict the future trajectory or sequence of control actions. Note that each vehicle can also aggregate information about other agents before taking the appropriate action. Using accurate state information rather than ego-vehicle mounted sensors such as RGB cameras has 2 additional advantages: 1) The computational burden on the resources can be relieved since images with many pixels being processed independently on each vehicle is no longer necessary. 2) The domain shift problem caused by the rendering of images in simulation not matching reality should no longer be a concern. This is because we are using the state (location, orientation etc.) of the vehicles as an abstraction to  represent information about them. Hence, with this abstraction it would be possible to train a model on one domain and test on another as we demonstrate in the Experiments. 

\begin{figure*}
    \centering
    \includegraphics[width=1.8\columnwidth]{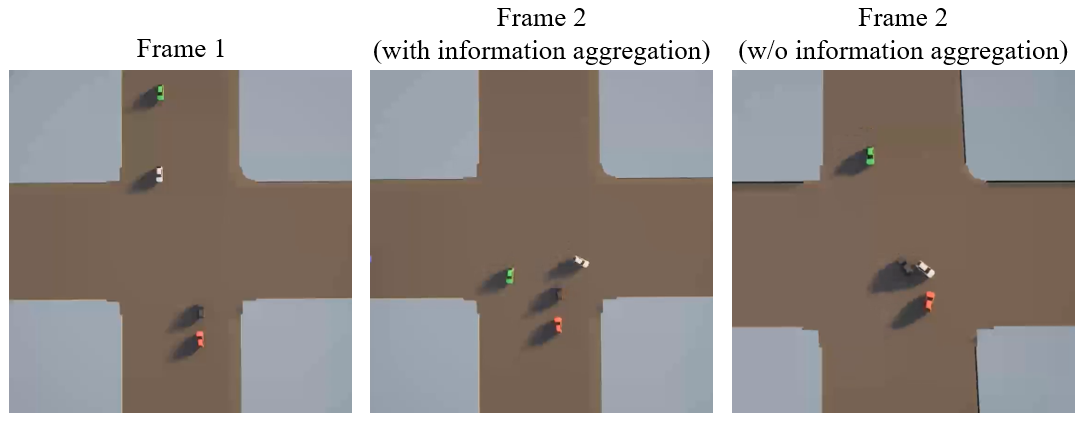}
    \caption{Figure depicts the importance of aggregating information. \emph{Left:} describes an initial scene comprising of 4 vehicles. The white vehicle is the first one arriving the intersection from the top and intends to turn to its left. Meanwhile, the brown and red vehicles arriving from the bottom intend to turn left and go straight respectively. \emph{Middle:} With information aggregation, the brown and red vehicles wait until the white vehicle has left the intersection. \emph{Right: } With no information aggregation, the brown and red vehicles start moving earlier and crash into the white vehicle.}
    \label{fig:sumo_teaser}
    \vspace{-4mm}
\end{figure*}

Figure \ref{fig:overview} further shows that the state information of all vehicles along with their desired intention to go straight, turn left or right is passed to a Multi-vehicle Trajectory Prediction (MTP) module. The MTP module predicts the future trajectory for each vehicle based on this provided information. Within the MTP, the future trajectory prediction is in turn done by the aggregation module which has shared weights across all the vehicles. This allows the model to handle an arbitrary number of vehicles in the scene. Note that to make a prediction for a particular vehicle, the aggregation module not only takes information about that specific vehicle but also considers information of other vehicles through message passing \cite{gilmer2017neural}. This provides each vehicle a holistic overview of the environment, thereby making an informed trajectory prediction. In contrast, Figure  \ref{fig:sumo_teaser} shows the implications of not aggregating information from other vehicles when making trajectory predictions. To this end, the contributions of this work are summarized below:

\begin{enumerate}
    \item We demonstrate that our approach of using only the state and intention information outperforms the approach of using past trajectory information. 
    \item Our model has the ability to predict the future trajectory of an arbitrary number of vehicles. It aggregates information from other vehicles; thereby giving better predictions. 
    \item We show that the model can be trained on one platform and tested on another. 
    \item Our approach of predicting the future trajectory can easily be extended to also control multiple vehicles simultaneously at intersections.
    \item We have also released the entire codebase for training and testing our method. The code can be found here: \url{https://github.com/Dekai21/Multi_Agent_Intersection}\color{red}.
\end{enumerate}
Note that the primary emphasis of this work is the MTP module in Step 3 of Figure \ref{fig:overview}, where we show how the future trajectory of multiple vehicles can be predicted simultaneously using their state and intent information. Step 1, regarding retrieving the BEV information of the vehicles and Step 2 regarding transmission of intention information to the MTP is touched upon in the related work Section \ref{sec:related_work}. Based on this, the following assumptions are made:
\begin{enumerate}
    \item Access to the state information and intention of the vehicles is available. 
    \item In case of control, the vehicles are capable of receiving the control commands to execute the correct maneuvers at intersections. 

\end{enumerate}

\section{Related Work}\label{sec:related_work}

\noindent{\textbf{Wireless Vehicle Communication:}}\\ Vehicle to Everything/Infrastructure (V2X/V2I) allows for wireless communication with the vehicles \cite{houmer2022secure}. V2X/V2I offers various advantages over the usual line of sight sensors placed on ego-vehicles \cite{7535520}. It facilitates reliable data transmission \cite{VASSEUR2010335}  even among non-line-of-sight vehicles in the immediate vicinity to give prior warnings of impending traffic jams \cite{8911628}, emergency braking \cite{ZHAO2004291}, risky overtaking \cite{8671732}. In our framework, it would be needed to transmit state information from and control commands to the vehicles. However, the emphasis of our work is multi-vehicle trajectory prediction and not vehicle communication. Therefore, we assume that the intention information is known by utilizing different trajectory datasets/platforms in our experiments.\\

\noindent{\textbf{Trajectory Datasets:}}\\ There are multiple real world datasets which provide trajectory information of various road participants from a BEV perspective. For e.g. \cite{Lerner2007CrowdsBE,5459260} collect data from top of buildings, while \cite{978-3-319-46484-8_33,inproceedingsTopview} collect from drone imagery. However, these datasets contain limited proportion of trajectories comprising of vehicles in the scene. This would not be enough in terms of quantity and accuracy to train data driven learning based algorithms. We therefore train and test on the real world inD dataset \cite{9304839}, which addresses these limitations. Note that our approach of trajectory prediction can also be extended to control multiple vehicles. This falls under the purview of embodied agent evaluation \cite{DBLP:journals/corr/abs-1807-06757} and is an emerging topic in the area of deep learning. However, none of the datasets described above provide the facility to conduct an online evaluation \cite{codevilla2018offline}. We therefore use the Simulation of Urban Mobility (SUMO) platform \cite{8569938}. In the context of this work, SUMO allows creation and control of various scenarios at intersection for e.g. the number/intention of vehicles, the priority of the roads, structure of the intersection etc. After training on SUMO, we then evaluate the online control of the vehicles on a completely different platform i.e. Car Learning to Act (CARLA) \cite{Dosovitskiy17}. CARLA provides the option to pass the steering and acceleration/throttle commands to maneuver multiple vehicles in the scene.\\ \\
\noindent{\textbf{Multi-agent trajectory prediction:}}\\
Future trajectory prediction of agents using information about the social interaction between them is being used for both pedestrians~\cite{TANG2022333,PENG2022258,ZHOU2021298,zheng2021unlimited} and vehicles~\cite{9700483,9564510,Zhao_2019_CVPR,8917228}. Many such methods utilize information about the past trajectory of vehicles to make inferences about the future \cite{vectornet,pmlr-v164-jia22a}.  In \cite{li2019interaction,ma2019wasserstein} the output is probabilistic, while being multimodal in \cite{sriram2020smart,9138768} particularly at points where a road splits into multiple directions. However, only one of the multiple alternatives would be valid if the vehicle intends to traverse a certain direction.  Our method in contrast does not require information about the past trajectory but rather only the current state of the vehicle. Also, the predictions of the future trajectory is unique as our model is conditioned on the intention of the vehicle. \cite{7353877} showed that being aware of the intention of other vehicles improves merging at T-intersections.   Knowledge of intention allows our task of trajectory prediction to be extended to additionally control the vehicles to reach desired targets. This is done by applying model predictive control to determine the appropriate throttle and steering angle such that the vehicle follows the predicted trajectory. We are not aware of any previous approaches that take only the current state and intention for future trajectory prediction and control of multiple vehicles. A recent work by \cite{9998111}, does use state and intent information but only for the task of maintaining a longitudinal safety distance between the front and rear vehicles. Moreover, their approach uses a rule based approach, whereas our approach is data-driven by training a neural network. \\ \\
\noindent{\textbf{Multi-agent Control:}}\\
Controlling a single vehicle at an intersections is a complicated task \cite{wang2021decision}. This is further aggravated when interaction with other agents also needs to be handled \cite{ge2021real}. The work of \cite{LIU2022390,9151648,8873518,8917268} control the flow of multiple vehicles to minimize traffic congestion and collision at intersections. However, this is done by controlling the traffic lights. Our network on the other hand deals with controlling the individual vehicles at intersections that are void of traffic lights. \cite{MOORTHY2022308,HE2022651} handle multiple agents using a leader guided formation control. In our work, all vehicles are independently controlled. Other approaches to control the individual vehicles involve solving an optimization problem \cite{levin2017conflict,kloock2019distributed}. Our approach in contrast is learning based.  \cite{WANG202068,palanisamy2020multi,ZHANG2021383,SIMOES202040} uses reinforcement learning (RL) for multi agent prediction/control. However, RL methods tend to be heavily data-inefficient \cite{electronics9091363}. Our framework, on the other hand uses imitation learning complemented with an additional collision cost to prevent vehicle-to-vehicle collision when controlling multiple vehicles simultaneously. 
\section{Framework}\label{sec:method}
In this section we describe the details of the Multi-Vehicle Trajectory (MTP) module depicted in Fig.\ref{fig:overview}. It takes the state and intention information of each of the $N$ vehicles in the scene as input and predicts their future trajectory for $T$ timesteps ahead.  We summarize the components of our framework as follows:\\

\noindent{\textbf{Input:}}\\ The information input to the MTP about each vehicle is represented by the vector ${X_k}$ $\in$ $\mathbb{R}^{6}$, $k = 1,2,...N$.   ${X_k}$ in turn comprises of 2 components: 1) State and 2) the intention of the vehicle.
\emph{State:} The state of vehicle \emph{k}  is in turn represented by a vector $\in$ $\mathbb{R}^{3}$, described by its orientation ($\theta_k$ $\in$ $\mathbb{R}$) and location ${S_k}$  $\in$ $\mathbb{R}^{2}$ on the $x-y$ plane. \emph{Intention:} of vehicle \emph{k}  represented by $I_k$ $\in$ $\mathbb{R}^{3}$ is a one hot encoded vector describing whether the vehicle intends to go either left, right or keep going straight at the upcoming intersection.\\

\noindent{\textbf{Input Transformation:}} \\ This input vector ${X_k}$ for each vehicle is then passed through a series of $L$ Multi-Layer Perceptron (MLP) layers with trainable parameters. The output of MLP layer $l$ for each vehicle $k$ is a latent representation given by ${X^l_k}$ $\in$ $\mathbb{R}^l$ and is specified by the following equation:

\makeatletter
\newcases{centercases}{\quad}
  {\hfil$\m@th\displaystyle{##}$\hfil}
  {$\m@th\displaystyle{##}$\hfil}{\lbrace}{.}
\makeatother

\begin{equation}
X^{l}_{k}=
\begin{centercases}
  X_{k}         & l=0 \\
  \sigma(\mathbf{W}^{l}{X}^{l-1}_k+\mathbf{b}^{l}) & 0<l \le L
\end{centercases}
\end{equation}

where $\mathbf{W}^{l}$ $\in$ $\mathbb{R}^{l \times (l-1)}$ and $\mathbf{b}^{l}$ $\in$ $\mathbb{R}^l$ are the trainable parameters of the MLP layer $l$, while $\sigma$ is the ReLU non-linear activation function.\\

\noindent{\textbf{Information Aggregation:}}\\ Note that the output of the last MLP layer $L$ for vehicle $k$ is ${X}^{L}_k$ and is only dependent on the latent representation of the same vehicle in the previous layer. In order to make an informed prediction of the future trajectory of a vehicle, it would be prudent to not only consider latent information about itself but also the other vehicles too. Therefore, information aggregation is done through message passing in the successive layers $l = L+1, L+2,...L_F$. This produces a new latent representation of each vehicle given by the following equation \cite{morris2019weisfeiler}:

\begin{equation}
\vspace{-1mm}
X^{l}_{k}=
\begin{centercases}
  X^{L}_{k}         & l=L \\
  \sigma(\mathbf{W}^{l}_s{X}^{l-1}_k+\mathbf{W}^{l}_o \sum_{p=1, p\neq k}^{N}{X}^{l-1}_p) & L<l < L_F \\
  \mathbf{W}^{l}_s{X}^{l-1}_k+\mathbf{W}^{l}_o \sum_{p=1, p\neq k}^{N}{X}^{l-1}_p & l = L_F
  \label{eq:aggregation}
\end{centercases}
\end{equation}

where $\mathbf{W}^{l}_s$, $\mathbf{W}^{l}_o$ $\in$  $\mathbb{R}^{l \times (l-1)}$ are the trainable parameters of the aggregation layer. The output of each vehicle $k$ in the last layer is $X^{L_F}_{k}$ $\in$ $\mathbb{R}^{2T}$. It is a prediction of the future trajectory information $S_k$  of the vehicle $k$ for T timesteps ahead. Note that in the experiments, we demonstrate the significance of aggregating information from neighbouring vehicles. Therein, we show that the performance of the trained model significantly deteriorates when the second term in Eq. \ref{eq:aggregation} corresponding to aggregation of information from the neighbouring nodes is removed.

Note that despite having shared weights, each vehicle predicts a unique trajectory, since the input vector given by the state and intention information for each vehicle is different. \\

\noindent{\textbf{Loss Function:}} \\
The loss function used to train our model can be decomposed into the imitation loss ($L_{imitation}$) and the collision loss ($L_{collision}$). The imitation loss is the mean of the $L_2$ distance between the the future trajectory predicted by the model and the ground truth.

\begin{equation}
\vspace{-1mm}
L_{imitation} = \frac{1}{N} \sum_{k=1}^{N} \sum_{t=1}^{T} {|S^t_k - \hat{S^t_k}|}_2
\end{equation}

where $S^t_k$ and $\hat{S^t_k}$ are respectively the predicted and ground truth state information of vehicle $k$ at timestep $t$.  Meanwhile, if the future trajectory of any 2 vehicles (e.g. vehicle $i$ and vehicle $j$) coincide within a certain safety distance threshold  $\lambda$ at the same time instance $t$, then a collision cost proportional to the excess is added as part of the collision loss: 
\begin{equation}
L_{collision} =  
  \sum_{i,j} {L_{collision_{i,j}}}
  \vspace{-3mm}
\end{equation}

\begin{equation}
L_{collision_{i,j}}=
\begin{centercases}
0         & if \min_{t} {{|{S^t_i} - {S^t_j}|}_2} > \lambda\\
\lambda - \min_{t} {{|{S^t_i} - {S^t_j}|}_2}         & otherwise
\end{centercases}
\label{eq:collision}
\end{equation}

where $1 \leq i < j \leq N$ and $1 \leq t \leq T$. 
The purpose of the collision loss is to mitigate the propensity of vehicle-to-vehicle collision at intersections. We demonstrate the importance of this component of the loss function in the experiments.\\ \\
\noindent{\textbf{Vehicle Control:}}\\
Note that our Multi-Vehicle Trajectory module can be extended to also control the individual vehicles. For this, we model the car with the bicycle model \cite{wang2001} and apply model predictive control (MPC) to optimize for the acceleration ($a$) and steering angle ($\delta$) such as to follow a selected $J$ number of points on the predicted trajectory of the vehicle. MPC has demonstrated to be of better performance compared to other controllers \cite{9517448,JRC8800}. The equation of motion considering the bicycle model are given by:

\begin{equation}
\vspace{-1mm}
    \dot{x} = v \cdot cos\theta;\\  ~\dot{y} = v \cdot sin\theta; ~\dot{v} = a; ~\dot{\theta} = v  \frac{\tan\delta}{L}
    \label{eq:motion_model}  
\end{equation}

where $L$ is the wheelbase and $v$ is the velocity of the vehicle. Meanwhile, the cost function minimized during optimization is given by:
\begin{equation}
    \min_{a,\delta} \sum_{i=1}^{J}  [ (x\textsubscript{i} -  \hat x_i)^2 + (y\textsubscript{i} -  \hat y_i)^2 +  (\theta\textsubscript{i} -  \hat \theta_i)^2]
    \label{eq:cost_fn}  
\end{equation}

\noindent{\textbf{Data Augmentation:}}\\
Note that when it comes to online vehicle control, training merely on the recorded data may not be enough. This is because the parameters controlling the car may cause the ego-vehicle to diverge from the expected trajectory. This deviation from the norm would cause the ego-vehicle to reach scenarios not seen by the model during training, such as the lane of the oncoming traffic or the road boundaries. Since, such scenarios are not present in the training set, the prediction of future trajectory by the model would be incorrect causing the control parameters to further deviate the car from the normal trajectory such that it eventually crashes into the side barrier. Therefore, to prevent these collisions with the barrier, we additionally augment the original recorded data by adding some noise to the position of the car. The output future trajectory is then determined using model predictive control described by Eq.  \ref{eq:motion_model} and \ref{eq:cost_fn}. However, the only difference is that, the optimization is to be done only for the final point on the trajectory, rather than on the $J$ points on the known trajectory. Experiments show that inclusion of this augmentation reduces collisions with the barrier during vehicle control.

\section{Experiments}
To measure the performance of our framework, we conduct both an offline and online evaluation. Offline evaluation is an assessment of future trajectory prediction of a trained model. For this we use the real world inD dataset \cite{9304839}.

Note that our approach of trajectory prediction is also capable of being extended to control the driving of individual vehicles at intersections. However, offline evaluation may not necessarily reflect the true driving quality. In fact, \cite{codevilla2018offline} showed that 2 models with similar offline metrics can have drastically different performance when deployed in a live setting.  For this, online evaluation wherein the agents can actively interact with the environment is necessary. Therefore, we use the CARLA \cite{Dosovitskiy17} platform for online evaluation with the model trained on a different platform i.e. SUMO \cite{SUMO2018}. The SUMO-CARLA co-simulation facilitates this evaluation.
 
\subsection{Offline Evaluation:}\label{subsec:offline}
\begin{figure}
    \centering
    \includegraphics[width=1\columnwidth]{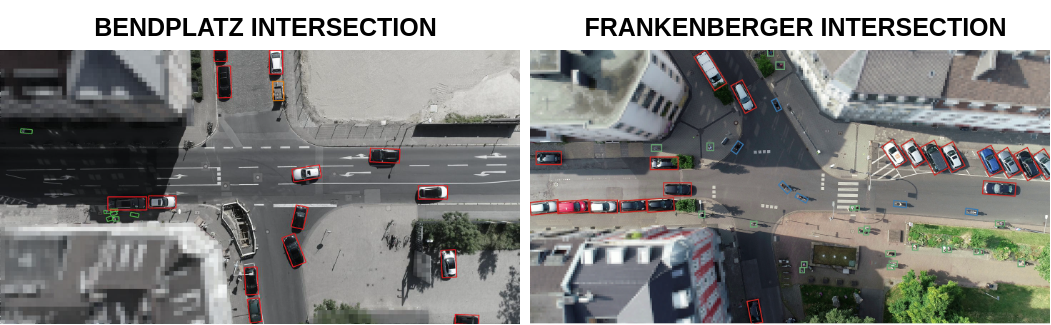}
    \caption{A snaphot of the Bendplatz and Frankenburg intersections in Germany available in the inD dataset \cite{9304839}}
    \label{fig:inD_intersections}
    \vspace{-3mm}
\end{figure}

The  Bendplatz  and  Frankenburg intersections from the inD dataset shown in Figure \ref{fig:inD_intersections} have been used for offline evaluation. For each intersection, 3 track files for training and 1 for validation are randomly selected. Each track file contains 20 minutes of track records collected during different times. 4 commonly used offline evaluation metrics are used for comparison, namely: Average Displacement Error (ADE) , Final Displacement Error (FDE), Miss Rate (MR) and Collision Rate (CR) \cite{vectornet,interactiondataset}.  For the interested reader, mathematical formulation and interpretation of these metrics, along with further information regarding the inD dataset used in the experiments is provided in the supplementary file\footnote{\url{https://github.com/Dekai21/Multi_Agent_Intersection/tree/master/supplementary}}.

 Apart from our model, 3 additional models were trained for purpose of comparison. Description of which are described below:\\
 
\noindent\textbf{{Past Trajectory (VectorNet):}}\\ This model is adapted from the approach of \cite{vectornet}. Like our approach, it retrieves information about the surrounding vehicles. It uses an attention based mechanism for this purpose.  However, this approach additionally uses information of not just the current state of the vehicle but also the past trajectory in order to better ascertain the future trajectory.  Our model in contrast  uses the intention of where the vehicle desires to go rather than the past trajectory. \\

\noindent\textbf{{No information aggregation:}} \\ The architecture of this model is similar to our model, except that a vehicle does not aggregate information from other vehicles in the environment. This is done by preventing message passing among the vehicles for trajectory prediction. Moreover, only the imitation learning loss is used for training.\\

\noindent\textbf{{No collision cost:}}\\ This is also similar to our approach, except that this model is trained without the additional collision cost we introduced into our imitation learning paradigm. 

\noindent\textbf{{Ours:}} \\ This model is trained using the framework described in Section \ref{sec:method}. The model takes information about the state and intention of each vehicle in the scene and predicts the future trajectory for each based on this available information. Note that the model is capable of making each vehicle aggregate information of other vehicles via message passing. This  holistic representation of the environment ought to facilitate an informed trajectory prediction that minimizes collisions between the multiple agents. The model is trained with both the imitation and collision loss functions. However, note that data augmentation meant for online vehicle control is not done here. 
 
The result of offline evaluation for all the 4 models are given in Table.\ref{ind1_offline} and Table.\ref{ind2_offline}.

\begin{table*}
\setlength\abovecaptionskip{-0.2cm}
\captionsetup{justification=centering}
\caption{\label{ind1_offline}Results of trajectory prediction at the Bendplatz Intersection in the InD Dataset. (Lower metric values are better)}
\begin{center}
\scriptsize
\scalebox{1.4}{
\begin{tabular}{ |c||c|c|c|c||c|c|c|c|c| } 
\hline
Model & \makecell{Past \\ Traj.} & \makecell{Intention} & \makecell{Message \\ Passing} & \makecell{Collision \\ Cost} & \makecell{ADE } & \makecell{FDE} & \makecell{MR} & \makecell{CR} & MR+CR\\
\hline
\hline
\makecell{Past \\ trajectory} & \checkmark & & \checkmark & & 3.800 & 7.515 & 0.816 & 0.043 & 0.859\\
\hline
\makecell{No info. \\aggregation} & & \checkmark & & & 1.341 & 2.619 & 0.230 & 0.127 & 0.357\\
\hline
\makecell{No\\collision cost} & & \checkmark & \checkmark & & 1.110 & 2.193 & 0.172 & 0.101 & 0.273 \\
\hline
\makecell{Our\\model} & & \checkmark & \checkmark & \checkmark & 1.099 & 2.126 & 0.157 & 0.075 & \textbf{0.232}\\
\hline
\end{tabular}
}
\end{center}
\end{table*}

\begin{table*}
\setlength\abovecaptionskip{-0.2cm}
\captionsetup{justification=centering}
\caption{\label{ind2_offline}Results of trajectory prediction at the Frankenburg Intersection in the InD Dataset. (Lower metric values are better)}
\scriptsize
\begin{center}
\scalebox{1.4}{
\begin{tabular}{ |c||c|c|c|c||c|c|c|c|c| } 
\hline
Model & \makecell{Past \\ Traj.} & Intention & \makecell{Message \\ Passing} & \makecell{Collision \\ Cost} & \makecell{ADE} & \makecell{FDE} & \makecell{MR} & \makecell{CR} & MR+CR\\
\hline
\hline
\makecell{Past\\trajectory} & \checkmark & & \checkmark & & 2.192 & 4.437 & 0.513 & 0.092 & 0.605\\
\hline
\makecell{No info.\\aggregation} & & \checkmark & & & 1.958 & 3.924 & 0.411 & 0.147 & 0.558\\
\hline
\makecell{No\\collision cost} & & \checkmark & \checkmark & & 1.752 & 3.518 & 0.341 & 0.112 & 0.453\\
\hline
\makecell{Our\\model} & & \checkmark & \checkmark & \checkmark & 1.850 & 3.623 & 0.359 & 0.072 & \textbf{0.431}\\
\hline
\end{tabular}
}
\end{center}
\vspace{-4mm}
\end{table*}

\subsection{Online Evaluation/Control}\label{subsec:online}
Online evaluation of the driving quality is done on the CARLA platform. However, the model is trained on data from SUMO. The intersection is created such that the vertical road (top-bottom) has higher priority over the horizontal road (left-right). The metric used for evaluation of online driving quality is the  \emph{Distance Collision Ratio (DCR)}. It is an online metric describing the distance covered by the agents before either a vehicle-to-vehicle (V2V) or vehicle-to-barrier (V2B) collision occurs.  It is mathematically described as the total distance driven by all the vehicles at an intersection over the total number of V2V or V2B collisions that occur. A higher value of this metric is better.\\
\begin{equation}
    DCR = \frac{1}{C} \sum_{k=1}^{N} \sum_{t=1}^{T_{k}-1} \sqrt{(x_{t+1,k} - x_{t,k})^{2} + (y_{t+1,k} - y_{t,k})^{2}}
\end{equation}
\newline
 where \emph{C} is either the number of V2V or V2B collisions. Meanwhile, $T_{k}$ is the number of timesteps it takes for a vehicle $k$ to cross an intersection. Generally, it is larger for vehicles taking a left turn as opposed to those taking a right turn due to the difference in the length of the circumference of the respective curvatures.\\ Models used for comparison in this online evaluation are the same as described in Subsection \ref{subsec:offline} for offline evaluation. The only difference is that 2 additional models are trained with data augmentation to enhance robustness to deviations caused by imprecise predictions. The first model is trained with data augmentation but no collision loss and the other model is trained with both augmentation and collision loss.\\  DCR metric for V2V and V2B collision for all these models are described in Table \ref{tab:carla_online}. 
For purpose of reproduciblity, the inference code for online control and the details of the SUMO-CARLA co-simulation setup are provided in the following repository: \url{https://github.com/Dekai21/Multi_Agent_Intersection#run-the-inference-code}.

\begin{table*}

\setlength\abovecaptionskip{-0.2cm}
\caption{\label{tab:carla_online}Results of Online Evaluation on CARLA. (Higher metric values are better)}
\begin{center}
\scriptsize
\scalebox{1.43}{
\begin{tabular}{ |c||c|c|c|c|c||c|c| } 
\hline
Model & \makecell{Past \\ Traj.} & Intention & \makecell{Message \\ Passing} & \makecell{Collision \\ Cost} & \makecell{MPC \\ Aug.} & \makecell{DCR (V2V)}& \makecell{DCR (V2B)}\\
\hline
\hline
\makecell{Past\\trajectory} & \checkmark & & \checkmark & & & 99.1 & 158.6\\
\hline
\makecell{No info.\\aggregation} & & \checkmark & & & & 168.7 & 607.4\\
\hline
\makecell{No collision cost \&\\augmentation} & & \checkmark & \checkmark & & & 722.2 & 515.9\\
\hline
\makecell{No\\augmentation} & & \checkmark & \checkmark & \checkmark & & 925.2 & 341.3\\
\hline
\makecell{No\\collision cost} & & \checkmark & \checkmark & & \checkmark & 753.6 & 1256.0\\
\hline
\makecell{Our\\model} & & \checkmark & \checkmark & \checkmark & \checkmark & \textbf{3915.0} & \textbf{1957.5}\\
\hline
\end{tabular}
}
\end{center}
\vspace{-5mm}
\end{table*}

\subsection{Discussion:}
In this subsection we elaborate some findings from the results. \\

\noindent{\textbf{Significance of aggregation:}}\\ As can be seen, the model with no aggregation of information from other vehicles under-performs our model. This is because, intersections are locations where plenty of interaction among multiple vehicles is expected to happen. Therefore, with no aggregation, an agent only receives information about itself and is oblivious to the state, intention and behaviour of the other vehicles. Hence, it cannot holistically look at the entire scene before taking an informed decision about its own trajectory prediction. In case of online evaluation on CARLA, we observed something interesting. Most crashes occurred not within the intersection but rather just before the vehicle enters the intersection on the non-priority road. This is because in the training set, these vehicles yield the right of way to those on the priority road by slowing down or even stopping completely before entering intersection. This is to allow the vehicles on the priority road to pass without hindrance. Only when there is no hindrance to other vehicles, the vehicle on the non-priority road moves in to the intersection. However, such situations are very rare compared to the number of samples where the vehicle on the non-priority road sits stationary. Hence, without receiving  knowledge about other agents, the model memorizes to always remain stationary before entering the intersection from the non-priority road. This blocks the non-priority road and prevents other vehicles from passing. In an ideal world, if a vehicle is blocking a road, the other vehicles approaching this choke point will be expected to slow down to prevent a crash. However, these vehicles are also oblivious to the presence of the blocking vehicle and attempt to drive through it causing a crash. This particularly lowers the DCR  metric especially for V2V collisions. 

\begin{figure}
    \centering
    \includegraphics[width=1\columnwidth]{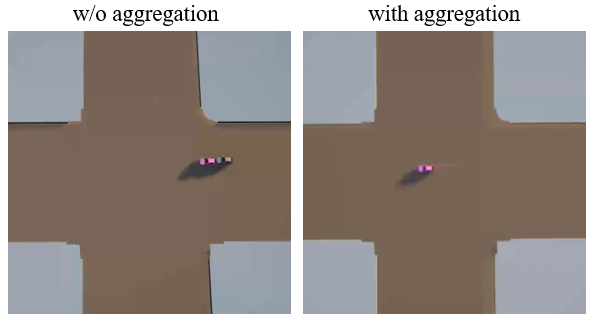}
    \caption{Shows an example of the implications of not using aggregation in comparison to our model which uses aggregation at an intersection on the CARLA simulator. The horizontal lane (left-right) is the non-priority road, while the vertical lane (top-bottom) is the priority road. }
    \label{fig:without_attention}
    \vspace{-4mm}
\end{figure}

Figure \ref{fig:without_attention} demonstrates the consequences of not aggregating information.  As described earlier, this leads to collisions among the vehicles before they enter the intersection due to the first stationary vehicle. On the right side of the same figure, an example scenario of our model which uses aggregation is presented. The pink vehicle desiring to go straight moves into the intersection as it is aware that there is no other vehicle at the intersection. \\

\noindent{\textbf{Contribution of Collision Cost:}}\\ It can be observed that our model which uses the collision cost penalty during training performs better than the model trained without it. The effect is even more pronounced  on the online metric particularly when it comes to preventing V2V collisions. Note that the DCR metric for V2V drops significantly when this loss component is removed from the training. The utility of the collision cost is that it has the ability to make slight modifications to correct the trajectory of the vehicles if it senses a potential collision thereby providing it with the ability to evade other vehicles.
The supplementary material contains a video demonstrating the implications when collision cost is not used as opposed to our approach.\\

\noindent{\textbf{Importance of Data Augmentation:}}\\ Note that we introduced data augmentation to prevent the vehicle from deviating and crashing into road barriers during online evaluation. Comparing the performance of the model trained without data augmentation shows that the DCR metric is significantly reduced particularly for V2B collisions. Our model in contrast was trained with data samples at deviated positions from the normal trajectory. Hence, even if the model were to end up at  divergent positions during online inference, it would know the corrective action to take to bring the vehicle back on track. This prevents crashes with the barrier or other vehicles if they are in the way. \\
\begin{figure*}[ht]
    \centering
    \includegraphics[width=1.9\columnwidth]{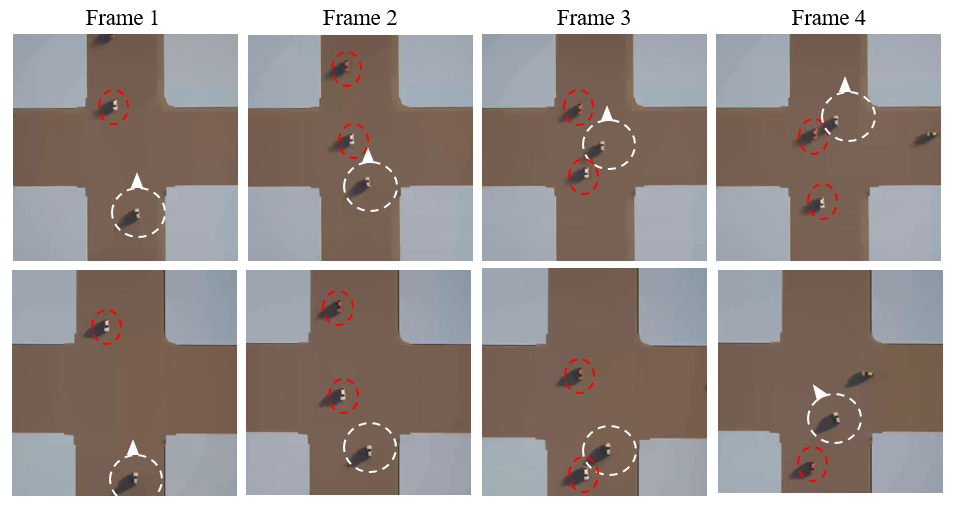}
    \caption{Demonstrates how intention can be used to control the behaviour and interaction among the vehicles. In the first row of images, the white circled vehicle coming from the bottom desires to go straight. It keeps moving without yielding to any other vehicle. In the second row, the intention of the same vehicle is modified to turn left. In this case, the white vehicle slows down to yield to the red circled vehicles which are moving straight. The white vehicle only starts executing the left turn once the red vehicles have passed. The arrow on the white circle represents the direction of motion. There is no arrow in case the white circled vehicle is stationary. Note that the vehicle coming from the right intends to turn right, so it is not a hindrance when the white circled vehicle intends to turn left. }
    \label{fig:change_intent}
    \vspace{-4mm}
\end{figure*}

\noindent{\textbf{Past Trajectory information:}}\\ Recall that the model in \cite{vectornet} uses past trajectory information of a vehicle in order to predict the future trajectory. Hence, such models have a probabilistic interpretation, wherein the precise future trajectory tends to be fuzzy and begins to become more precise by the time the vehicle reaches well into the intersection. In contrast, since our model is provided with information about the intention of the vehicle, the predictions are unique and much more accurate as can be seen from the results. This intention allows our approach to be extended to vehicle control.  Figure \ref{fig:change_intent} shows that by manipulating the intention, the interaction among the vehicles is adjusted accordingly. This flexibility in changing the behaviour is only possible due to the capability derived from using intention of the vehicle at the input. Not only are the offline trajectory predictions more accurate (see Table \ref{ind1_offline} and \ref{ind2_offline} ) but the online control is also more robust (see Table \ref{tab:carla_online}) in comparison to using past trajectory information.\\

\noindent{\textbf{Domain Adaptation:}}\\ Note that our model trained only on data from the SUMO platform to predict the future trajectory can also be used to control the vehicle on a completely different platform. In this case, it is the CARLA platform. Note that data from CARLA was not available to the model during training. The reason for this successful adaptation of the model to different domains is because we are using the state information of the vehicle as the representation. This representation remains consistent across different platforms/domains. Hence, the model is immune to the source of origin of this representation i.e. CARLA or SUMO. Other representations such as images have difficulty in switching between different domains, weather/lighting conditions etc. For e.g. a control model trained on images from a sunny weather condition would have difficulty controlling the vehicle in a rainy weather condition even though the domains may be the same \cite{control-across-weathers-19}.

Note that the entire code for training and conducting both offline along with online evaluation is contained in the following repository: \url{https://github.com/Dekai21/Multi_Agent_Intersection}.

\section{Conclusion}

In this paper, we demonstrated how the trajectory for multiple vehicles can be predicted simultaneously at intersections. This is done by utilizing their state and intention information. This allowed extending the approach to additionally controlling the vehicles to move towards desired directions. Aggregating information of other vehicles further facilitated each vehicle to make better informed decisions. Our framework is also capable of being trained on one domain while being tested on another domain, data of which was not seen during training.\\

\bibliographystyle{IEEEbst} 
\bibliography{root.bib}

\end{document}